# Good Things Come in Trees: Emotion and Context Aware Behaviour Trees for Ethical Robotic Decision-Making

Paige Tuttösí[1], Zhitian Zhang[1], Emma Hughson[1], Angelica Lim[1]

1. Simon Fraser University, School of Computing Science

**Executive Summary**

Emotions guide our decision making process and yet have been little explored in practical ethical decision making scenarios. In this challenge, we explore emotions and how they can influence ethical decision making in a home robot context: which fetch requests should a robot execute, and why or why not? We discuss, in particular, two aspects of emotion: (1) somatic markers: objects to be retrieved are tagged as negative (*dangerous*, e.g. knives or *mind-altering*, e.g. medicine with overdose potential), providing a quick heuristic for where to focus attention to avoid the classic Frame Problem of artificial intelligence, (2) emotion inference: users' valence and arousal levels are taken into account in defining how and when a robot should respond to a human's requests, e.g. to carefully consider giving dangerous items to users experiencing intense emotions. Our emotion-based approach builds a foundation for the primary consideration of Safety, and is complemented by policies that support overriding based on Context (e.g. age of user, allergies) and Privacy (e.g. administrator settings). Transparency is another key aspect of our solution. Our solution is defined using behaviour trees, towards an implementable design that can provide reasoning information in real-time.

**Our Solution**

Introduction to the Ethical Context

Robots are here to improve our lives in many ways; they can be employed to assist those with mobility, sight or hearing impairments, or even simply as an assistant to perform mundane tasks and free up our time (Riek, 2017). With these advancements and the overall novelty of introducing robots into our daily routines, the ethical implications and considerations of household robotic assistants' actions are often left behind.

Many of the tasks we assign to a household robot would appear simple to a human and require little effort in the decision making process, however, our brains have the ability to process moral and ethical decisions with amazing speed and efficiency. Our so-called "moral intuition" (Haidt, 2001) quickly draws upon our learned cultural, social, emotional and situational knowledge and makes simple decisions such as "Should I give this 5 year old this pocket knife?" almost instantly without much conscious effort. Yet, robots have no inherent ethical decision making properties; these must be provided by those who design them. We can ask a robot to not provide users under a certain age a dangerous object, but what defines a dangerous object? What should the age cut off be? In what situation may it be necessary that they have a dangerous object? What frequency of requests for dangerous objects should become questionable?

Ethical decision-making frameworks suggest that people evaluate situations using a combination of moral intuition and moral reasoning (Schwartz, 2016). Social interactionist perspectives of morality suggest that consequences of an action are considered before deciding whether an action is a moral violation (Turiel, 1998). Even from the young age of 3, children are able to judge right from wrong, based on whether an action is harmful or helpful (Dahl, 2018). Emotions, importantly, play a role in moral judgement and its processing (Greene et al., 2001). Emotions have been suggested to have evolved to address decision-making in the Frame Problem (De Sousa, 1987), a classic issue examined in artificial intelligence, where agents have difficulty deciding which facts about the world are relevant or irrelevant. To avoid spending unbounded amounts of time examining all possible consequences of a decision, emotions serve as a heuristic to quickly know what can be ignored as irrelevant, and to focus our attention. On the other side of the coin, emotions have also been thought to be an antithesis to rational decision making (Solomon, 1993). Emotional states can influence decision making, and intense negative emotions can disrupt self-control (Pham, 2007). The varied influences of emotion within ethical decision making are ripe for further investigation.

Ethical Design Considerations

We have been tasked with designing an ethical policy for a household assistant robot that fetches objects for the users. There are no limitations on the physical abilities of the robot and we have applied assumptions that the robot has several contextual sensors.

In designing policies for household robots, balancing ethical decisions with usefulness and generalisability is a challenge. We have three key stakeholders: (1) The user, (2) The developer, (3) The company; each with competing goals for ethical implementation. The user wants a robot that is safe, trustworthy and useful. The developer wants policies that are easy to implement and that can generalise to all citations. The company wants this solution to be innovative, adaptive and reproducible in many contexts. How do we ensure that our robot is safe yet not become trapped in a situation where we limit the possible actions to a degree that the robot is no longer useful. Additionally, how do we ensure that this policy is both generalisable and implementable so that it can easily be produced at a commercialisable scale. In order to address all stakeholders, we made several key considerations in developing our policies keeping in mind key consideration from Riek & Howard, 2014 (Appendix):

- Safety: Err on the side of caution. At the end of the day the average user can simply fetch the object themselves, they are inconvenienced, but not placed in a more dire situation than if the robot were not present.
- Transparency: We want both users and designers who implement the design to be able to clearly understand why a decision has been made. We have chosen a tree based design so that policies can be logically followed.
- Privacy: We use administrator roles to protect personal objects.
- Implementability : The design method should be easy to compose, modify and implement. We choose behaviour trees as our design, and it can be implemented on various platforms. There are two open-source behaviour tree libraries that are available: PyTree[1] and BehaviorTree.CPP[2]. Both are compatible with the ROS Ecosystem.
- Generalisability : By providing broad groupings we hope to encompass as many possible users and objects into our policies as possible. Those implementing this system should, hopefully, be able to find a place in the tree for any encountered object or user.

Policies

We have 3 overarching policies: 1. **Object ordering (safety) policy**, 2. **Emotional policy**, 3. **Object Category policy**; and in some cases: 4. **Personal (privacy) policy**, 5. **Context policy**. All policy checks must be completed in order for the object to be accessible to the user.

Our decision making charts can be found in the Appendix (pg. 7-10). Each bright green box performs a "Policy Check," with red representing a danger safety branch, orange representing a substance safety branch and yellow designating a personal/privacy branch. Following the tree, the resulting leaves will be either pink, indicating that no groups may receive the requested object, or green, listing the groups that are allowed to receive the object. If the user is identified as a member of this group they will be provided the object. All groups, aside from unknown, include: adult (over 19, or 21 in the USA, designated A), teen (aged 13-18, designated T), and child (aged 5-12, designated C). All those under the age of 5 may not receive objects from the robot. The groups include: household (HA, HT, HC)[3], friends (FRA, FRT, FRC), family (FAA, FAT, FAC)[4], and unknown (U).

We begin with the **Object Ordering Policy**: the system will maintain memory of the last object requested by the user. If a *dangerous object*[5] is selected, the 30 minute cool-down period will begin. If a *mind altering object*[6] is selected, a four hour cool-down period will begin. If in either of these cool-down periods, depending on the emotional context, several branches become unavailable. For example, no objects in the "vehicle" category can be given when a mind

---

[1] https://py-trees.readthedocs.io/en/devel/index.html
[2] https://www.behaviortree.dev/
[3] Household can be set by the admin and may include those who lime in the home, but also others who visit on a regular basis and may need access to more sensitive objects such as cleaning staff or assistants
[4] Family members not residing at the residence
[5] Objects that are clearly known to cause harm to oneself, or others, or other objects when used for the intended purpose or if misused or improperly handled
[6]Includes possible non mild altering medicines but with overdose potential. This group is for substances meant for consumption, other possibly dangerous objects if improperly ingested would be 'dangerous'

altering object is on cool-down. For all other branches **a Context Policy** check, i.e. sensors (for example sensing the room the user is in) and a verbal response from the user must support that the object is being used for the object's intended task, and the resulting "allowed" user groups are reduced. If the user is denied an object that is either dangerous or mind altering, the current cool-down period is reset to reflect the cool-down of the denied object. These contextual checks of dangerous objects and their interaction with previously requested objects reflect somatic markers (Damasio, 1996) that humans place on negatively viewed objects.

Next is an **Emotional Policy** check where we borrowed from the idea of emotional valence and arousal. Valence indicates level of positive or negative emotion (Shuman., Sander, & Scherer, 2013), while arousal indicates the level of physical reaction to a given emotional state (Deckert, Schmoeger, Auff & Willinger, 2020). Using a grid (Kollias & Zafeiriou 2020), we place a user's current emotion on a x- and y- axis to map out their current state (figure 1.). Certain areas on this grid are considered dangerous, concerning/hazardous, cautious, or safe. Emotions associated with a specific zone are provided and indicate where they are placed in terms of arousal and valence. The dangerous category are those in the *red zone* of the graph, which indicate highly fragile states or well-being, mostly associated with extreme anger. If in the *red zone* and requesting a dangerous or mind altering object in most cases all groups are denied as this could lead to physical or emotional damages. Emotional states with high levels of arousal, except for sadness, are associated with impairments in working memory capacity (Pham, 2006) and could impact sound reasoning. In the *orange zone* we have our category of concern and require more **Context Policy** checks and limits dangerous and mind altering objects (especially when previously coupled with mind altering objects). In this area we have extreme sadness, such as depression, and as such certain objects, like mind-altering or dangerous objects ones, include additional barriers of access as we want to reduce the risk of harm (e.g., suicide). In the *yellow zone* is an area of caution and requires additional **Context Policy** checks. However, this zone is more flexible and less severe than the *orange* or *red zone*. This zone is more for situations where someone is not in a positive or neutral mood (i.e., high valence and high/low arousal) but they don't have extreme negative moods (i.e., high/low arousal and low valence).

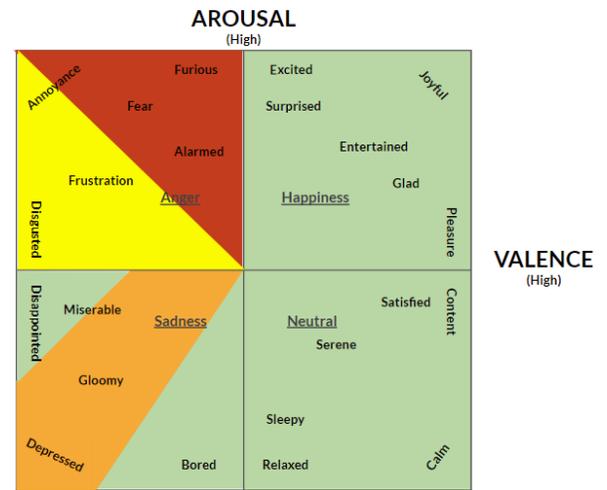

Figure 1. Emotional zone chart

Lastly, an **Object Category Policy** check will be implemented. Depending on the object category several **Context Policy** checks may be required and can include: age, presence of an adult (for children), intended use and even allergies. These checks will often be verbal affirmations but may be aided by sensors, e.g. sensing the presence of an adult or room context. These are the most specific policy checks and must consider the entirety of the previously followed tree when designing the final decision. Depending on the object class and the branch followed there may be a **Personal Policy** check. This is handled by admin settings. The admin settings are hierarchical where the owner of the policy makes the first selection on who can designate personal objects. Objects previously designated as personal may not be selected by the next additional user (additional users are defined by the policy owner). This policy protects objects as important as medication and personal identification, but also as simple as your own personal clothing.

<u>Behaviour trees</u>

Behaviour trees are hierarchical control systems which are similar to hierarchical finite state machines. Behaviour trees (BTs) are designed to be traversed in a specific order, whether a simple check or a complex action, until a terminal state is achieved. The major benefit of using BTs is its modularity and reactivity (Colledanchise & Ögren, 2018). These two properties make the system that is designed with BTs to be efficient and explainable.

In this work, we adopt behaviour trees for our novel ethical design. As shown in Figure 2, the root of our tree is a decorator node which modifies its child node with a custom policy. In this case the decorator's policy is repeat, the tree will repeat itself for every item that the robots are asked to fetch. Next in the tree is a sequence node which executes its children nodes in order. In our design, the first executable child node is for the robot to check if it has any required knowledge of the request such as previous requested item, identity of the human, emotion of the human and context information of the environment. If there is no previous knowledge, the robot will try to extract that information under the assumption that the robot has the required sensors and algorithms to obtain that information, and save it to the blackboard. A blackboard is a storage place for a behaviour tree where individual behaviours can read or write data. Otherwise, the next child node will be executed to read previous information from blackboard as well as update the current information. After the robot acquired the necessary information, the next child node is a series of hierarchical fallback nodes. The fallback node works in a way that it will check if there is any violation on the policies we designed. If a violation exists, the tree will move on to the next execution node which will return a decline decision for the request. The fetch request will only be accepted if there is no violation for all five policies we designed.

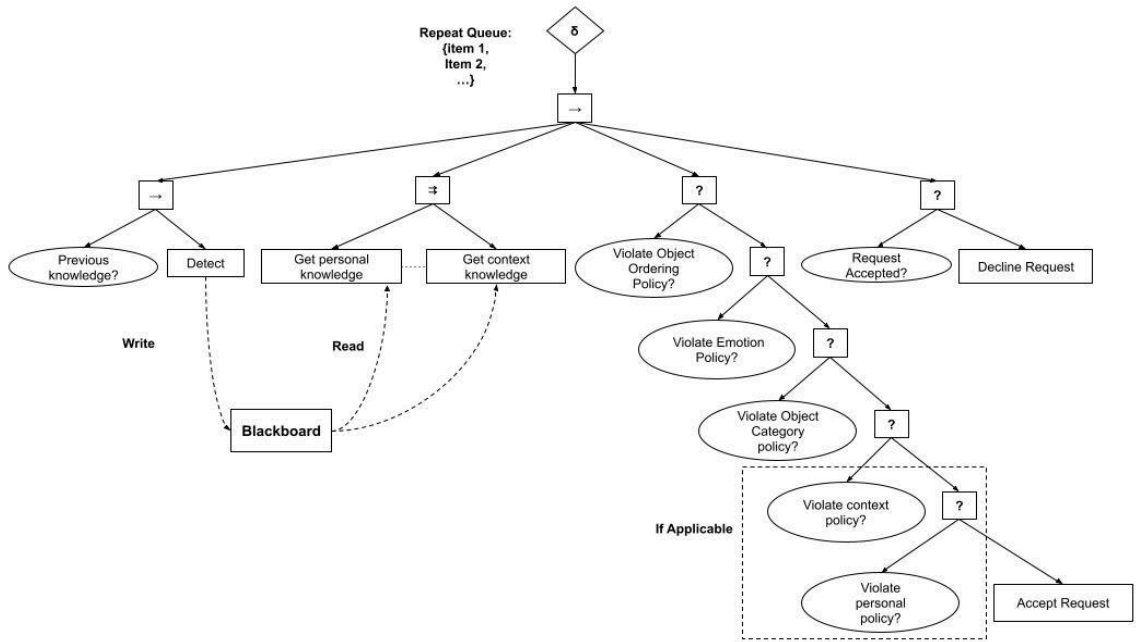

Figure 2. Behaviour Tree Design

Impact

Our proposed solution allows for generalizability across several different contexts but also includes user affect to make decisions. Given the close ties between emotions and decisions it is with no surprise that emotions should be incorporated when making ethical choices. Emotions are what make us human and can also cause harm. Therefore emotions are at the centre of our problem space. Although robots do not currently have the capability to feel/understand emotions, our use of arousal and valence gives robots a space to map human affect.

Evaluation and Limitations

While privacy and safety policies are relatively straightforward, the addition of tracking of objects and cool-down time for "dangerous" objects helps to improve the intricacy of both of these policies mimicking human's emotionally controlled negative somatic markers. Furthermore, our novel emotional policy implementing valence and arousal scales adds a layer of context detection attempting to replicate on a simple scale some of the ways humans employ their "moral compass" through emotional heuristics.

Although we have addressed policies of safety, privacy, legality and emotions, one particularly difficult problem to solve is making moral decisions. These decisions may not necessarily be dangerous or illegal, but do have negative moral implications. We hope that both emotional and context policies can begin to allow robots to make these types of decisions, however, more research needs to be done specifically to improve this area of ethical robotic decision making.

Testing

Our design solution can be tested in both simulations and in real life. In simulation, we can provide test scenarios with information that are required by our behaviour trees and design policies. Our solution could be implemented with the open source library mentioned earlier and output results for the information given. In real life, our behaviour tree can be implemented easily on any robot with ROS platform. And as long as the robots are given high level information of the task, emotions, item category etc, our solution could be tested for ethical decision making.

**Appendix**

We highlight key considerations from The Code of Ethics for the Human-Robot Interaction Profession (Riek & Howard, 2014) to include when creating interactive household robots:

*Human Dignity Considerations*
(a) The emotional needs of humans are always to be respected.
(b) The human's right to privacy shall always be respected to the greatest extent consistent with reasonable design objectives.

*Design Considerations*
(d) Maximal, reasonable transparency in the programming of robotic systems is required.
(e) Predictability in robotic behaviour is desirable.
(f) Trustworthy system design principles are required across all aspects of a robot's operation, for both hardware and software design, and for any data processing on or off the platform.

*Legal Considerations*
(i) All relevant laws and regulations concerning individuals' rights and protections (e.g., FDA, HIPPA, and FTC) are to be respected.
(j) A robot's decision paths must be re-constructible for the purposes of litigation and dispute resolution.

*Social Considerations*
(o) Avoid racist, sexist, and ableist morphologies and behaviours in robot design.

Chart Examples

Top Level Chart:

**Chart A: Previous Object and Emotion Policies**

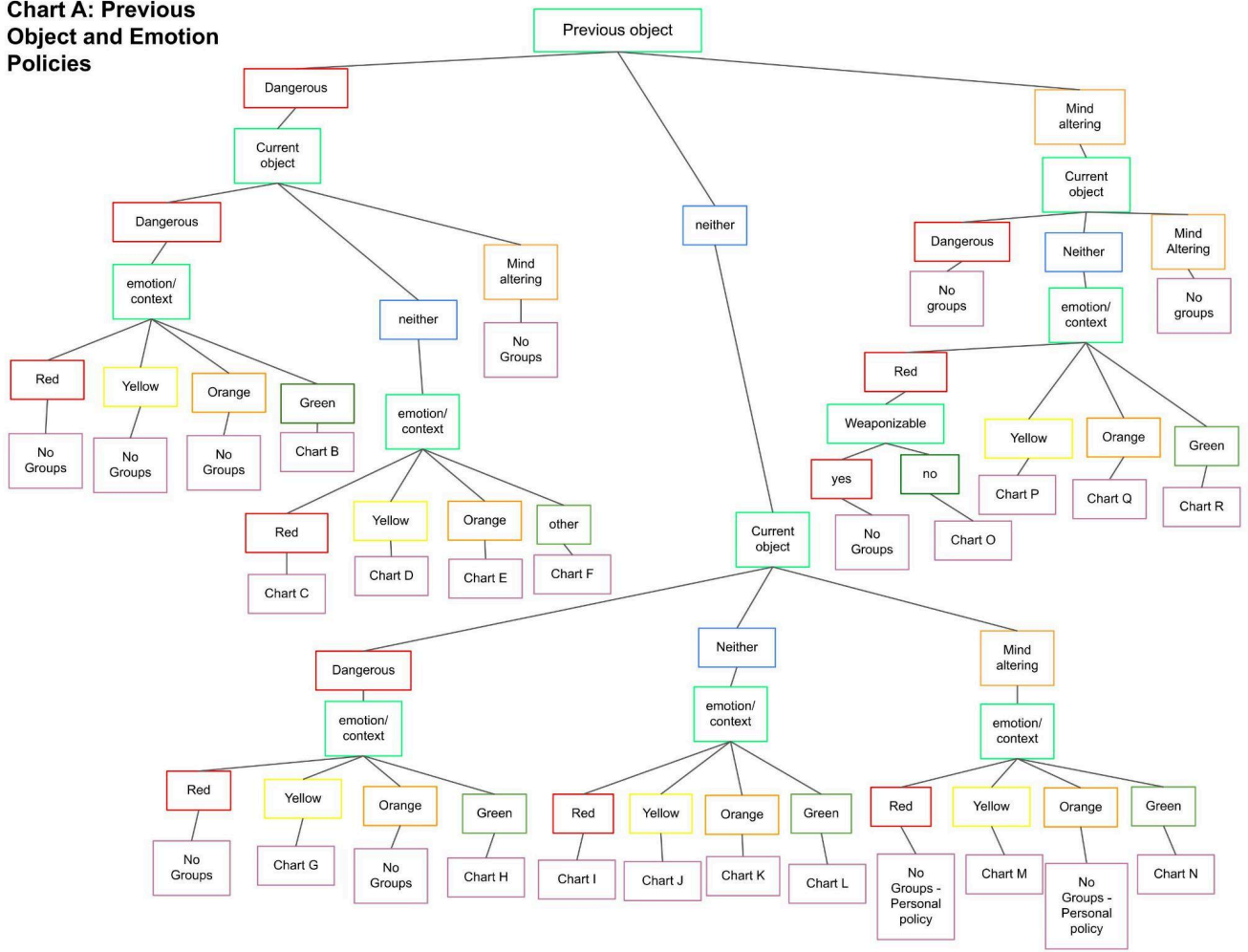

Danger Branch example:

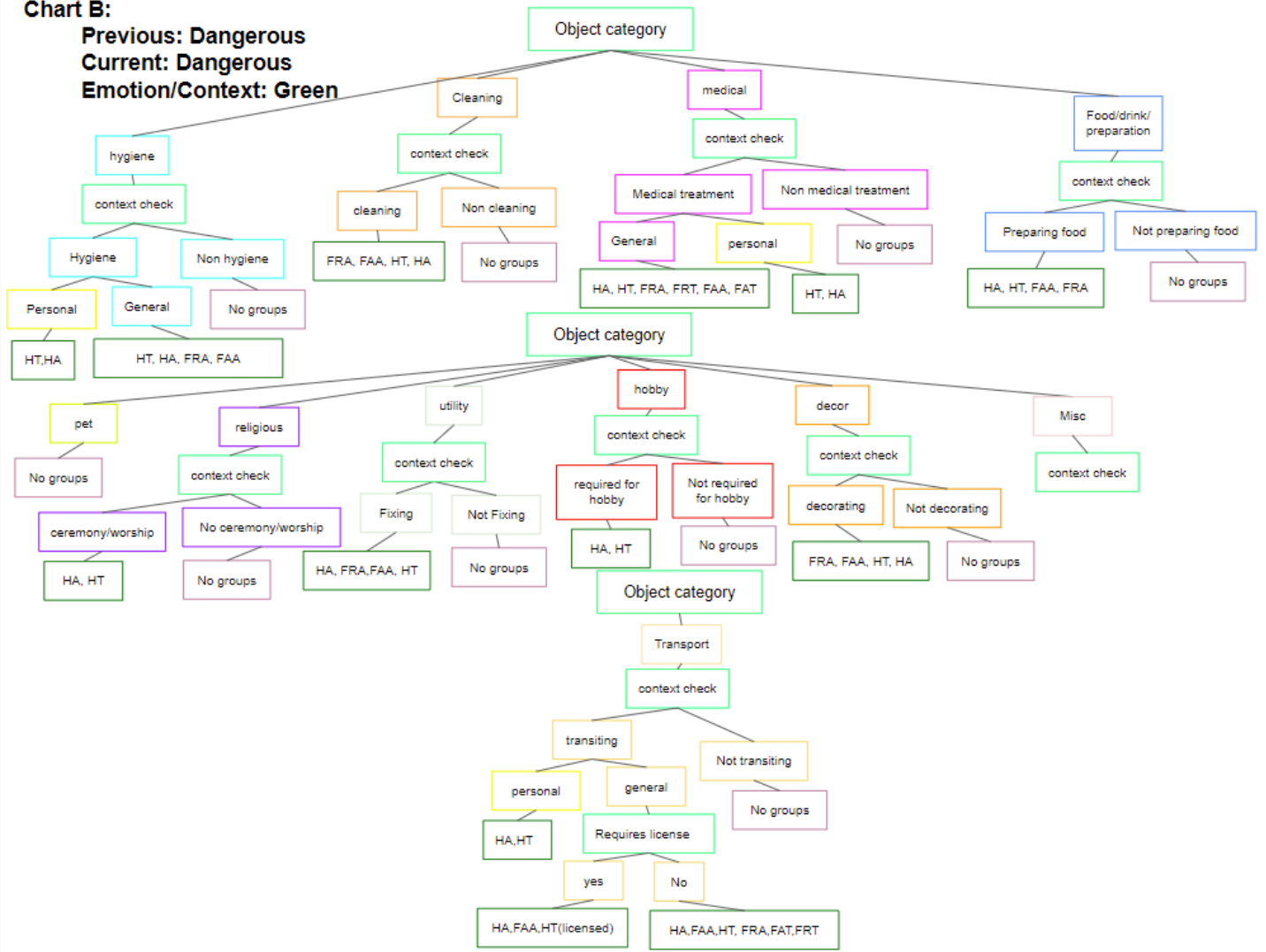

Mind altering example:

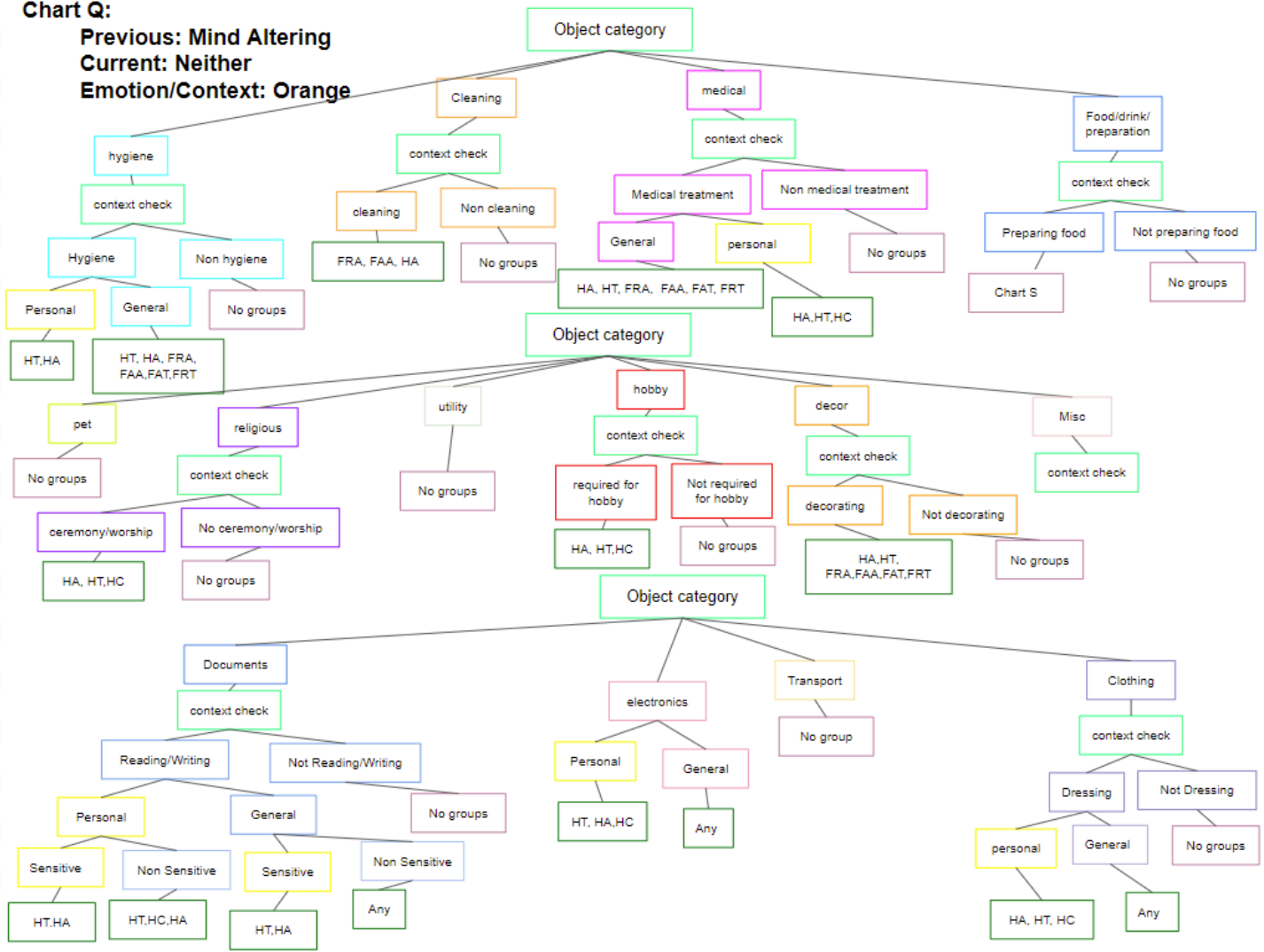

Example of object levels when all danger and emotional policies are passes in the green zone:

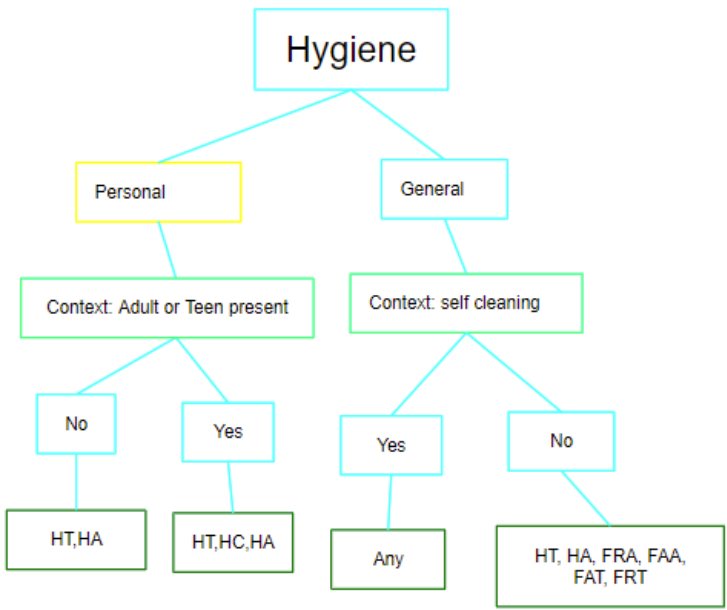

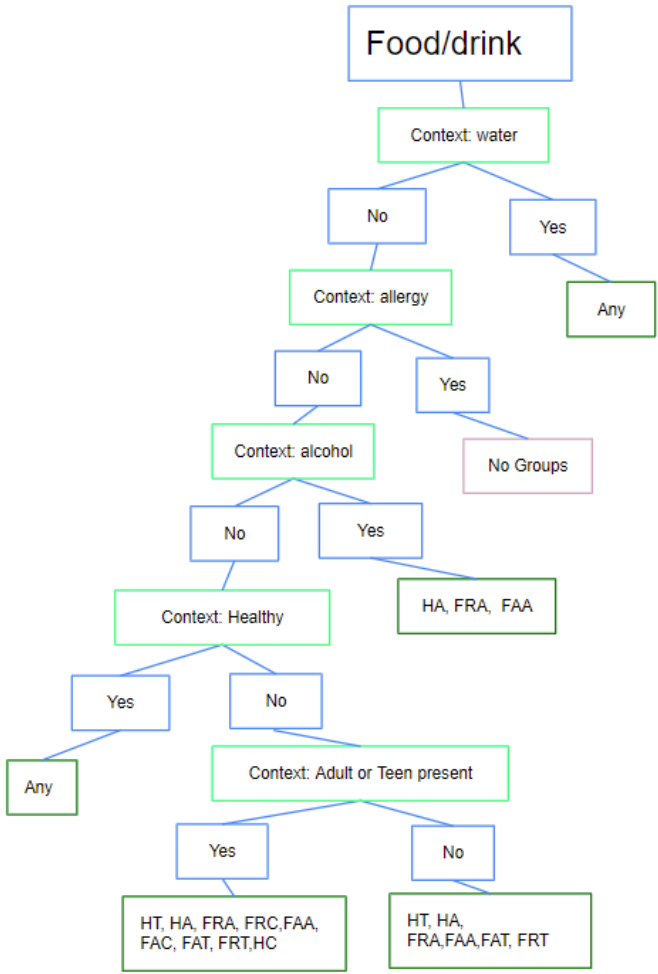